\documentclass[
]{ceurart}

\usepackage{listings}
\usepackage{import}
\usepackage{tikz}
\usepackage{pgfplots}

\begin{document}

\copyrightyear{2022}
\copyrightclause{Copyright for this paper by its authors.
  Use permitted under Creative Commons License Attribution 4.0
  International (CC BY 4.0).}

\conference{Deep Learning for Knowledge Graphs (DL4KG) 2023}

\title{Universal Preprocessing Operators\\for Embedding Knowledge Graphs with Literals}

\author[1]{Patryk Preisner}[%
orcid=0000-0001-6950-4459,
email=patryk.konrad.preisner@students.uni-mannheim.de,
]
%\cormark[1]
\address[1]{Data and Web Science Group, University of Mannheim, Germany}

\author[1]{Heiko Paulheim}[%
orcid=0000-0003-4386-8195,
email=heiko@informatik.uni-mannheim.de,
url=http://www.heikopaulheim.com/,
]
\cormark[1]

\cortext[1]{Corresponding author.}
%\fntext[1]{These authors contributed equally.}

\begin{abstract}
Knowledge graph embeddings are dense numerical representations of entities in a knowledge graph (KG). While the majority of approaches concentrate only on relational information, i.e., relations between entities, fewer approaches exist which also take information about literal values (e.g., textual descriptions or numerical information) into account. Those which exist are typically tailored towards a particular modality of literal and a particular embedding method. In this paper, we propose a set of universal preprocessing operators which can be used to transform KGs with literals for numerical, temporal, textual, and image information, so that the transformed KGs can be embedded with any method. The results on the kgbench dataset with three different embedding methods show promising results.
\end{abstract}

\begin{keywords}
  Knowledge Graph \sep
  Embedding \sep
  Representation \sep
  Literal Information
\end{keywords}

\maketitle

\section{Introduction}
Knowledge graphs have become a common means to represent information across various domains. \cite{heist2020knowledge,hogan2021knowledge} They are comprised of entities and their relations, but many also contain literal information, like textual descriptions of entities, numerical values, or even images. For example, the following is an excerpt of the representation of the entity \emph{Mannheim} in DBpedia~\cite{auer2007dbpedia}:
\begin{verbatim}
dbr:Mannheim dbo:country dbr:Germany .
dbr:University_of_Mannheim dbp:city dbr:Mannheim .
dbr:Mannheim dbo:populationMetro "2362046"^^xsd:nonNegativeInteger .
dbr:Mannheim dbo:foundingDate "1607-01-24"^^xsd:date .
dbr:Mannheim dbo:abstract "Mannheim [...] officially the University City of 
   Mannheim (German: Universitätsstadt Mannheim), is the second-largest city 
   in the German state of Baden-Württemberg..."@en .
dbr:Mannheim foaf:depiction 
  <http://commons.wikimedia.org/wiki/Special:FilePath/
   NUB_Mannheim_2014-03-13.jpg> .
\end{verbatim}
Most embedding approaches only consider relations between entities when computing numeric representations for entities. In the above example, when learning a representation for the entity \emph{Mannheim}, they would use only the first two statements, but neglect the latter three, containing textual, numerical, and image information. However, those also contain relevant information about the entity, which could lead to a better latent representation if they were used by the embedding approach.

While a few embedding approaches have been proposed which take into account literal information, they have a few shortcomings: most of them (1) target only one modality (e.g., text, numbers, \emph{or} images), and (2) are adaptations of a particular embedding method and hence cannot be used in conjunction with arbitrary embedding methods.

In this paper, we propose a set of knowledge graph preprocessing operators for textual, numeric, and image literals which can be used to create a KG with only relations from one containing literal information. The resulting knowledge graph can then be processed by any arbitrary embedding method.

The rest of this paper is structured as follows. Section~\ref{sec:related} positions our approach in the light of existing research. Section~\ref{sec:approach} introduces our approach, followed by a set of experiments described in section~\ref{sec:experiments}. We conclude with a summary and an outlook on future work.

\section{Related Work}
\label{sec:related}
Many standard benchmarks for knowledge graph embeddings, especially in the link prediction field, do not come with literals. Hence, the topic has not drawn as much attention as knowledge graph embeddings for purely relational KGs for quite some time. 

A survey from 2021~\cite{gesese2021survey} lists a number of approaches, which mostly are extensions of existing knowledge graph embedding models, mostly classic models like TransE. Those approaches usually change the loss function of the underlying model and hence are bound to that model alone. An exception is LiteralE~\cite{kristiadi2019incorporating}, which has been applied to different embedding algorithms like TransE, ComplEx, or DistMult. Moreover, most approaches focus only on one modality of literals. 
A more recent survey from 2023~\cite{fanourakis2023knowledge} confirms that picture.

In contrast, the work presented in this paper proposes to preprocess a KG with literals in a way that the information in the literals is represented in a KG with only relational information. We investigate a number of preprocessing techniques for various modalities, which can be applied together with arbitrary embedding models.

The pyRDF2vec~\cite{pyrdf2vec} implementation of RDF2vec~\cite{ristoski2016rdf2vec} has a functionality to extract literals directly as features. This creates a heterogeneous representation of an entity (consisting of an embedding plus an additional vector of literal values), which is similar to the \emph{Data Properties} strategy described in~\cite{paulheim2012unsupervised}. In contrast, the approch in this work targets a uniform embedding representation.

An alternative is to alter the knowledge graph upfront, aiming at transforming information in encoded literals into relational statements. Such approaches would not be bound to a particular embedding method, and, if developed for literals with different modalities, could also be combined to exploit 
However, approaches based on preprocessing are still rare. One exception is~\cite{wang2022augmenting}, who propose the use of binning of numerical values. We reuse some of their approaches in our work in this paper. Another paper~\cite{blum2022exploring} also proposes three strategies preprocessing literals, one of which is used as a baseline in this paper.

\section{Approach}
\label{sec:approach}
\begin{figure}[t]
\caption{Overall Framework}
\label{fig:framework}
    \centering
    \def\svgwidth{\textwidth}
\footnotesize
        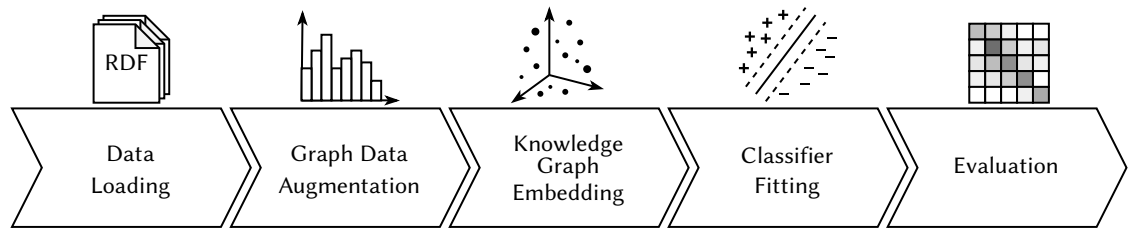
\end{figure}

Our approach relies on graph preprocessing. Instead of changing the embedding approach per se, we augment the graph with additional nodes and edges encoding some of the information encoded in the literals. Fig.~\ref{fig:framework} shows the overall framework. Specifically, the embedding step is decoupled from the augmentation step. The last two steps (classifier fitting and evaluation) are concerned with evaluation. For the experiments in this paper, we consider node classification problems, but other downstream tasks (such as link prediction, node regression, or node clustering) would also be possible.

\subsection{Baselines}
For all approaches, we employ three simple baselines. The first, tagged \texttt{EXCLUDE}, simply excludes all literals. Since most embedding approaches ignore literals, this should not have an impact.

The second, tagged \texttt{TRANSFORM}, creates an entity for each combination of a literal value and a property. In the example above,
\begin{verbatim}
dbr:Mannheim dbo:populationMetro "2362046"^^xsd:nonNegativeInteger .
\end{verbatim}
would be transformed to\footnote{Note that all of the approaches technically turn an \texttt{owl:DatatypeProperty} into an \texttt{owl:ObjectProperty}. If this is not wanted, e.g., since the ontology should be further reused, this can trivially be changed, e.g., by moving the property into a different namespace.}
\begin{verbatim}
dbr:Mannheim dbo:populationMetro new:populationMetro2362046 .
\end{verbatim}
This strategy is identical with the method called \emph{Literal2Entity} in~\cite{blum2022exploring}.

The third and final baseline, tagged \texttt{ONEENTITY}, creates one single entity for each relation. The idea is to capture any information that is indicated only by the presence or absence of a datatype property (such as \texttt{dbo:populationMetro}), regardless of the actual literal value, similarly to the \texttt{relation} strategy in~\cite{paulheim2012unsupervised}. This strategy would transform the above triple to
\begin{verbatim}
dbr:Mannheim dbo:populationMetro new:populationMetroAnyValue .
\end{verbatim}

\subsection{Handling Numeric Literals}
Creating a single entity for each literal value may not be a good strategy for capturing the semantics of that value. Besides scalability issues, two very similar literal values are indistinguishable from two very dissimilar ones. To counter those issues, we employ a number of additional techniques for representing numeric literals, based on binning.

\begin{figure}[t]
    \caption{Illustration of the \texttt{nBINS} Approach}
\label{fig:bins_processed}
    \centering
    \def\svgwidth{0.75\textwidth}
    \begin{scriptsize} 
        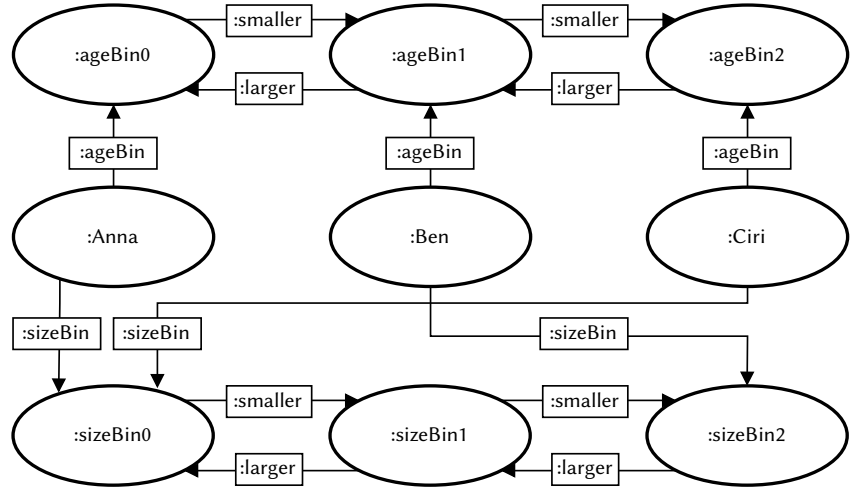
    \end{scriptsize}
\end{figure}

The most basic one, tagged \texttt{nBINS}, is simlar to the one proposed in \cite{wang2022augmenting}. We create $n$ bins from the set of literal values for each predicate. Furthermore, the entities representing the bins are connected to each other. Fig.~\ref{fig:bins_processed} shows the idea of this approach.
While \texttt{nBINS} requires setting a fixed value for $n$, \texttt{p\%BINS} lets the user set a percentage of unique values. For example, for a datatype property with 1,000 occurences, and 200 unique values, \texttt{10\%BINS} would create 20 bins (10\% of 200). Moreover, we also adapt the idea of \emph{overlapping bins} and \emph{hierarchical binning} from \cite{wang2022augmenting}, which allows for literal values to be contained in more than one bin, and therefore extends the expressivitiy of the entities representing bins.
%TODO NAMES for approaches
%TODO include figures if there still is space.

Since outliers can distort the bins created, we also combine the binning with a preceding outlier detection step. Specifically, we use the local outlier factor (LOF) method \cite{breunig2000lof} to first discard outliers, then perform a binning.

Finally, we adopt an idea from \cite{DetectError}, which is based on the observation that the same property may be used for multiple types of objects, hence resulting in different blended value distributions. For example, the property \texttt{height} may be used for people and buildings, but binning should be conducted on values from both classes separately, since the bin \texttt{high} would have a different span for people and buildings.

Since many knowledge graphs do not come with an extensive type system, we alter the original approach in~\cite{DetectError} to use either sets of relations for identifying similar and dissimilar entity types (in the example above, people and buildings would come with different sets of relations), and sets of relations and entities. The two approaches are coined \texttt{KL-REL} and \texttt{KL-RELENT}. Both approaches build a lattice of entities with the datatype at hand, and compute the KL-divergence of the set of relations (or the set of relations and the connected entities, respectively) and split the population of values until it falls below a certain threshold (in our experiments, we use 300 values as a threshold). Then, the binning is performed individually for each subpopulation.

All the approaches create one entity per bin and relation. Hence, the poulation statement in our example would be transformed to a statement like
\begin{verbatim}
dbr:Mannheim dbo:populationMetro new:populationMetroBin02 .
\end{verbatim}

\subsection{Handling Temporal Literals}
For temporal literals, i.e., literals typed with \texttt{xsd:date}, we follow a different strategy. The first strategy for handling dates, coined \texttt{DATBIN}, turns the date into a UNIX timestamp and applies the \texttt{nBINS} strategy above. In the above example, the statement
\begin{verbatim}
dbr:Mannheim dbo:foundingDate "1607-01-24"^^xsd:date .   
\end{verbatim}
would be replaced by a statement like
\begin{verbatim}
dbr:Mannheim dbo:foundingDate new:foundingDateBin14 .   
\end{verbatim}

This strategy, however, does not capture the entire information in a date. For example, a similarity of two people with the same birthday (in different years) might not be captured with such an approach. Therefore, to handle temporal literals, we propose a second strategy coined \texttt{DATFEAT} and extract five new features from a date literal. 

\begin{figure}[t]
    \caption{Date Nodes Encoding Quarter of Date}
    \label{fig:date_quarter}
    \centering
    \def\svgwidth{0.9\textwidth}
    \begin{scriptsize} 
        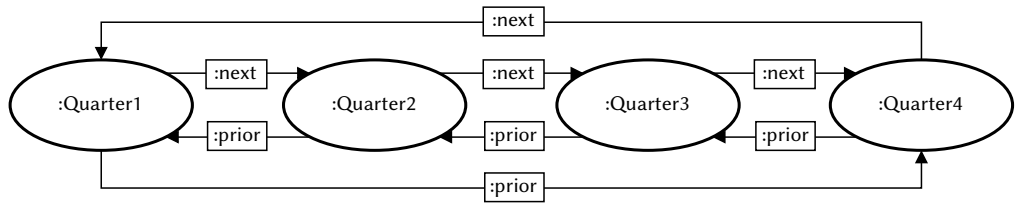
    \end{scriptsize}
\end{figure}

In the above example, this would yield the statements
\begin{verbatim}
dbr:Mannheim dbo:foundingDate
    new:wednesday ,
    new:day24 ,
    new:month1 ,
    new:quarter1 ,
    new:year1607 .
\end{verbatim}
As shown in Fig.~\ref{fig:date_quarter}, the new entities for days, months, and quarters can again be connected in order to also capture interrelations between them.

\subsection{Handling Text Literals}
Many knowledge graphs contain rich textual information, but this cannot be represented as easily as the information in numbers and dates. In order to represent textual information, we use \emph{topic modeling}, which assigns each text literal a certain number of topics~\cite{blei2003latent}. Each of those topics is then represented as a node in the graph.

Specifically, we run all values of a text literal (e.g., \texttt{dbo:abstract}) through a Latent Dirichlet Allocation (LDA) algorithm, and connect each entity to all topics exceeding a certain threshold (in our experiments in this paper, we use a threshold of 10\%). With this strategy coined \texttt{TXTLDA}, the statement
\begin{verbatim}
dbr:Mannheim dbo:abstract "Mannheim [...] officially the University City of 
   Mannheim (German: Universitätsstadt Mannheim), is the second-largest city 
   in the German state of Baden-Württemberg..."@en .
\end{verbatim}
could be replaced, e.g., by
\begin{verbatim}
dbr:Mannheim dbo:abstract 
    new:abstractTopic04, new:abstractTopic17 .
\end{verbatim}

\subsection{Handling Image Literals}
For images, we use a similar technique. We reuse a large-scale neural image classification model, which predicts tags for images (e.g., whether the building is showing a person or an animal). Those are then represented as nodes, which are then used to describe the image contents.

In our experiments, we use the pre-trained VGG16 model~\cite{simonyan2014very}, which computes probabilities for 1,000 classes of images. For each image, we classify it with VGG16 and use the most likely class for each image. In our example above, the triple
\begin{verbatim}
dbr:Mannheim foaf:depiction 
  <http://commons.wikimedia.org/wiki/Special:FilePath/
   NUB_Mannheim_2014-03-13.jpg> .
\end{verbatim}
could be replaced by
\begin{verbatim}
dbr:Mannheim foaf:depiction new:VGG_building .
\end{verbatim}

\begin{table}[t]
    \centering
    \begin{tabular}{l|c|c}
         Strategy &  $\delta E$& $\delta S$ \\
         \hline
         EXCLUDE & -- & -- \\
         TRANSFORM & V * R & S \\
         ONEENTITY & R & S \\
         nBINS & n*R & S \\
         DATBIN & n*R & S \\
         DATFEAT & DW+DD+DM+DQ+DY & 5*S \\
         LDA & T & T*S \\
         VGG16 & 1,000 & S \\
         \hline
    \end{tabular}
    \caption{Maximum size changes to the knowledge graph in number of entities ($\delta E)$ and statements ($\delta S$). Variables used: number of distinct literal values ($V$), number of relations ($R$), number of literal assignment statements ($S$), number of distinct weekdays ($DW$), days ($DD$), months ($DM$), quarters, ($DQ)$, and years ($DY$), topics in LDA ($T$).}
    \label{tab:strategies}
\end{table}
Table~\ref{tab:strategies} depictes the size changes of a knowledge graph for the individual strategies. It can be observed that the number of statements equals the number of original literal statements, and the number of entities is also changing only moderately.

\section{Experiments}
\label{sec:experiments}
We test all of the approaches above on the node classification benchmark \texttt{kgbench}~\cite{bloem2021kgbench}, which contains four heterogeneous datasets, as shown in table~\ref{tab:kgbench}. As embedding methods, we use TransE~\cite{bordes2013translating} and DistMult~\cite{yang2014embedding} using the pyKeen library~\cite{ali2021pykeen}, and RDF2vec~\cite{rdf2vecbook} using the pyRDF2vec library~\cite{pyrdf2vec}. As classifiers, we use kNN and SVM using the scikit-learn library~\cite{sklearn_api}.

\begin{table}[t]
\centering
\begin{tabular}{l|r|r|r|r}
     Dataset & amplus & dmgfull & dmg777k & mdgenre \\
     \hline
     Classes & 8 & 14 & 5 & 12 \\
     Relations & 33 & 62 & 60 & 154 \\
     Nodes & 1,153,679 & 842,550 & 341,270 & 349,344 \\
     Triples & 2,521,046 & 1,850,451 & 777,124 & 1,252,247 \\
     \hline
     objects thereof... & & & & \\
     ...IRIs & 1,464,871 & 593,291 & 288,379 & 1,001,791 \\
     ...blank nodes & 256,515 & -- & -- & -- \\
     ...literals & 799,660 & 1,257,160 & 488,745 & 250,456\\
     \hline
     thereof... & & & & \\
     ...numbers & 160,959 & 88,168 & 10,706 & 14,352 \\
     ...dates & 202,304 & -- & -- & 113,463 \\
     ...text & 377,542 & 834,244 & 329,987 & 54,838 \\
     ...images & 58,855 & 58,846 & 46,108 & 67,804 \\
     ...others & -- & 275,902 & 101,944 & -- \\
\end{tabular}
\label{tab:kgbench}
\caption{The kgbench dataset}
\end{table}

Using the Adam optimizer, the two pyKeen embedders DistMult and TransE were trained in 100 epochs for TransE and 150 epochs for DistMult, using the LCWA train loop. We use a batch size of 75,000 for DistMult and 2,000 for TransE. For all additional parameters, the default parameters provided by pykeen were used. Hereby the pykeen selects the parameters used in the original paper that introduced the selected embedder as default parameters \cite{ali2021pykeen}. RDF2vec was trained using a maximum walk depth and 500 walks per node, and 50 training epochs for word2vec. For all additional parameters, the default parameters of pyRDF2vec are used.
%\textbf{TODO no. of dimensions for all approaches}

For the clasifiers, we use a grid search for parameter optimization. For kNN, the parameters in the search space are $k = \{2,4,7,9,15\}$, for SVM, the parameters in the search space are $C = \{0.01, 0.1, 1, 10, 100\}$. For all other parameters, we use the default values defined by scikit-learn \cite{sklearn_api}.\footnote{The code for all experiments is available online at \url{https://gitlab.com/patryk.preisner/mkga/}}

Table~\ref{tab:results} shows the experiment results. For each literal type, we show the ones which got the best results overall, in addition to the three baselines.\footnote{A full table with the results for all configurations can be found at \url{https://gitlab.com/patryk.preisner/mkga/}.} Theses are KL-REL with LOF for numeric literals, DATBIN for dates (however, only amplus and mdgenre contain dates), LDA for text, and VGG16 for images. Moreover, we report results of a combined approach using the combination of the five aforementioned strategies.

From the table, we can observe that in three out of four cases, the best baseline can be outperformed by a few percentage points (0.779 vs. 0.708 on amplus, 0.676 vs. 0.606 on dmg777k, 0.726 vs. 0.662 on dmgfull), whereas for mdgenre, none of the approaches yields an advantage over the best baseline excluding literals (RDF2vec+SVM). 

Moreover, we can observe that there is no clear correlation between the amount of literals of a particular modality (see table~\ref{tab:kgbench}) and the improvement achieved by including the corresponding literals. While this might seem counter intuitive, the sheer amount of literals does not reflect the utility of the information contained therein.\footnote{As a thought experiment, imagine a numerical ID for each entity, which would greatly increase the number of numerical literals, but the literals would not contain any useful information.}

The baselines TRANSFORM and ONENENTITY are often strong competitors as well, indicating that in many of the cases, the presence of a literal is a strong signal, regardless of the actual literal value.

\begin{table}[t]
\centering
\caption{Experiment Results. The best results per dataset and embedding method are printed in bold, the best overall results per dataset are additionally underlined.}
\label{tab:results}
\footnotesize
\begin{tabular}{ll|rr|rr|rr|rr}
\toprule
 &  & \multicolumn{2}{c}{amplus} & \multicolumn{2}{c}{dmg777k} & \multicolumn{2}{c}{dmgfull} & \multicolumn{2}{c}{mdgenre} \\
 \cmidrule(lr){3-4}\cmidrule(lr){5-6}\cmidrule(lr){7-8}\cmidrule(lr){9-10}
 & & KNN & SVM & KNN & SVM & KNN & SVM & KNN & SVM \\
\midrule
\multirow[c]{8}{*}{\rotatebox[origin=c]{90}{DistMult}} & EXCLUDE & 0.458 & 0.512 & 0.548 & 0.542 & 0.619 & 0.658 & 0.605 & 0.622 \\
 & TRANSFORM & 0.477 & 0.546 & 0.593 & \textbf{0.611} & 0.560 & 0.576 & 0.575 & 0.610 \\
 & ONEENTITY & 0.511 & 0.549 & 0.517 & 0.501 & 0.613 & 0.649 & 0.599 & 0.617 \\
 & KL-REL+LOF & 0.564 & \textbf{0.608} & 0.528 & 0.523 & 0.634 & \textbf{0.673} & 0.616 & \textbf{0.632} \\
 & DATBIN & 0.501 & 0.538 & - & - & - & - & 0.609 & 0.626 \\
 & LDA & 0.464 & 0.504 & 0.564 & 0.582 & 0.652 & 0.665 & 0.612 & 0.623 \\
 & VGG16 & 0.485 & 0.528 & 0.549 & 0.552 & 0.553 & 0.642 & 0.606 & 0.621 \\
 & COMBINED & 0.542 & 0.590 & 0.579 & 0.583 & 0.595 & 0.643 & 0.612 & 0.620 \\
\hline
\multirow[c]{8}{*}{\rotatebox[origin=c]{90}{RDF2Vec}} & EXCLUDE & 0.550 & 0.536 & 0.586 & 0.606 & 0.629 & 0.661 & 0.590 & \textbf{\underline{0.662}} \\
 & TRANSFORM & 0.616 & 0.616 & 0.626 & 0.628 & 0.635 & 0.685 & 0.583 & 0.658 \\
 & ONEENTITY & 0.588 & 0.612 & 0.630 & 0.609 & 0.631 & 0.679 & 0.565 & 0.657 \\
 & KL-REL+LOF & 0.536 & 0.564 & 0.594 & 0.594 & 0.626 & 0.660 & 0.584 & \textbf{\underline{0.662}} \\
 & DATBIN & 0.554 & 0.523 & - & - & - & - & 0.591 & \textbf{\underline{0.662}} \\
 & LDA & 0.606 & 0.610 & 0.620 & 0.636 & 0.631 & 0.663 & 0.591 & 0.660 \\
 & VGG16 & 0.584 & 0.575 & 0.623 & \textbf{\underline{0.676}} & 0.627 & 0.651 & 0.591 & 0.660 \\
 & COMBINED & 0.688 & \textbf{0.691} & 0.664 & 0.665 & 0.650 & \textbf{0.715} & 0.586 & 0.661 \\
\hline
\multirow[c]{8}{*}{\rotatebox[origin=c]{90}{TransE}} & EXCLUDE & 0.682 & 0.708 & 0.506 & 0.528 & 0.649 & 0.662 & 0.634 & 0.646 \\
 & TRANSFORM & 0.727 & 0.761 & 0.602 & 0.611 & 0.665 & 0.688 & 0.639 & \textbf{0.649} \\
 & ONEENTITY & 0.737 & 0.761 & 0.610 & 0.621 & 0.657 & 0.673 & 0.641 & 0.647 \\
 & KL-REL+LOF & 0.716 & 0.726 & 0.476 & 0.512 & 0.643 & 0.664 & 0.638 & 0.644 \\
 & DATBIN & 0.683 & 0.719 & - & - & - & - & 0.634 & 0.642 \\
 & LDA & 0.670 & 0.701 & 0.554 & 0.578 & 0.653 & 0.672 & 0.631 & 0.648 \\
 & VGG16 & 0.723 & 0.735 & 0.560 & 0.588 & 0.656 & 0.671 & 0.632 & 0.644 \\
 & COMBINED & 0.760 & \textbf{\underline{0.779}} & 0.627 & \textbf{0.643} & 0.709 & \textbf{\underline{0.726}} & 0.635 & 0.646 \\
\bottomrule
\end{tabular}
\end{table}

\section{Conclusion and Future Work}
\label{sec:conclusion}
We have shown that graph preprocessing is a promising strategy for representing literal information in knowledge graph embeddings, which can be combined with arbitrary embedding methods.

The set of preprocessing operators is not fixed, but can be extended. For example, for text or image representation, while we used basic models to demonstrate the effectiveness of our approach, newer representation models can also be easily plugged in. A staged approach would also be feasible, e.g., representing texts first by means of a BERT encoder and then binning the resulting dimensional values.

Most of the approaches used do not only create entities (e.g., for numerical bins, topics, or image labels), but also come with some score for those. For example, LDA assigns probabilities to topics, given a text. In the experiments in this paper, we used a simple thresholding mechanism to include and exclude the corresponding edges, but it would also be possible to pass the scores to the embedding model as edge weights.~\cite{cochez2017biased}

%Overall, we have shown a flexible alternative to the numerous methods which adapt single embedding methods and are therefore bound to a limited number of modalities and embedding methods.

%\begin{acknowledgments}
  %Thanks to the developers of ACM consolidated LaTeX styles
  %\url{https://github.com/borisveytsman/acmart} and to the developers
  %of Elsevier updated \LaTeX{} templates
  %\url{https://www.ctan.org/tex-archive/macros/latex/contrib/els-cas-templates}.  
%\end{acknowledgments}

%%
%% Define the bibliography file to be used
\bibliography{sources_thesis_copy,sources_add}
\end{document}